\documentclass[conference]{IEEEtran}
\IEEEoverridecommandlockouts
\usepackage{cite}
\usepackage{amsmath,amssymb,amsfonts}
\usepackage{algorithmic}
\usepackage{graphicx}
\usepackage{textcomp}
\usepackage{xcolor}
\usepackage{float}
\usepackage{hyperref}    % hyperlinks
\usepackage{url}      % simple URL typesetting
\def\BibTeX{{\rm B\kern-.05em{\sc i\kern-.025em b}\kern-.08em
  T\kern-.1667em\lower.7ex\hbox{E}\kern-.125emX}}
  
\newcommand{\makefig}[3]{ \begin{figure#3}[h]\centering\includegraphics[width=\linewidth]{#1}\caption{#2}\label{fig:#1}\end{figure#3}}

\begin{document}
         
\title{BLM-17m: A Large-Scale Dataset for \\Black Lives Matter Topic Detection on Twitter}

\author{\IEEEauthorblockN{Hasan Kemik}
\IEEEauthorblockA{\textit{Information Technologies} \\
\textit{Carbon Consulting}\\
Istanbul, Turkey \\
hasan.kemik@carbonconsulting.com}
\and
\IEEEauthorblockN{Nusret \"Ozate\c{s}}
\IEEEauthorblockA{\,\,\,\,\,\,\,\,\,\,\,\,\,\,\,\,\,\,\,\,\,\,\,\,\textit{Department of Computer Science and Engineering\,\,\,\,\,\,\,\,\,\,\,\,\,\,\,\,\,\,\,\,\,\,\,\,} \\
\textit{Koc University}\\
Istanbul, Turkey \\
mozates23@ku.edu.tr}
\and
\IEEEauthorblockN{Meysam Asgari-Chenaghlou}
\IEEEauthorblockA{\textit{Research And Development} \\
\textit{Ultimate}\\
Berlin, Germany \\
meysam.asgarichenaghlou@ultimate.ai}
\and
\IEEEauthorblockN{Yang Li}
\IEEEauthorblockA{\textit{School of Automation} \\
\textit{Northwestern Polytechnical University}\\
Xi'an, China\\
liyangnpu@nwpu.edu.cn}
\and
\IEEEauthorblockN{Erik Cambria}
\IEEEauthorblockA{\textit{School of Computer Science and Engineering} \\
\textit{Nanyang Technological University}\\
Singapore, Singapore \\
cambria@ntu.edu.sg}
}

\maketitle

\begin{abstract}
Protection of human rights is one of the most important problems of the modern world. In this paper, we construct a Twitter dataset that covers one of the most significant human rights contradiction in recent years which affected the whole world: the \texttt{George Floyd} incident. We propose a labeled dataset for topic detection that contains about 17 million tweets. These Tweets are collected from 25 May 2020 to 21 August 2020, covering about 90 days from the start of the incident. We labeled the dataset by monitoring most trending news topics from global and local newspapers and used TF-IDF and LDA as baselines. We evaluated the results of these two methods with three different k values for precision, recall and F1-score.
\end{abstract}


\section{Introduction}
The George Floyd incident has affected many people of color and defenders of liberty and democracy. People protested this horrible incident for months around the USA and all over the planet. Despite the corona virus outbreak in months before the incident, many kept their faith in democracy and liberty by protesting. The importance of this incident has motivated many around the world to start researching many aspects of it~\cite{oriola2020covid,dreyer2020death,abc_george_floyd}.
%Social media on the other hand, plays a crucial role in many aspects of our lives and is an inseparable part of our daily life. The impact of Twitter, as a news platform and first witness reporting media on many events is undeniable. Such events can include sports, politics, protests, global news and many other topics. Easiness of sharing and receiving information across the globe within seconds also creates an environment for mass producing of data about the subject. 
%Despite the recent issues created by Elon Musk's takeover, Twitter still represents the number one platform for social media mining~\cite{clement_2020,castillo2011information,petrovic2013can}.
The hashtag \#BlackLivesMatter or \#BLM first emerged in 2013 following the acquittal of George Zimmerman in the shooting death of unarmed Black teen Trayvon Martin (Fig.~\ref{fig:blm}).
As the scourge of police brutality failed to subside over the years, the movement evolved and strengthened. The BLM movement sustained its momentum and activism after the 2014 shooting of Michael Brown by a white police officer.
In a span of just three weeks following the grand jury's refusal to indict the officer, the \#BlackLivesMatter hashtag was used over 1.7 million times on Twitter.
Finally, the hashtag culminated in 2020 with the George Floyd incident~\cite{clement_2020,castillo2011information,petrovic2013can}. 
%In~\cite{castillo2011information,petrovic2013can} Twitter's credibility and influence on reaching global audience is examined. Hence, with these properties, Twitter becomes an important platform for text and multimodal (image, text and video) data analytical practices. 
To have a better understanding of the social stance for the people concerned, supporting, or even in contradiction with the George Floyd incident, we collected a dataset containing about 17 million tweets. 

\makefig{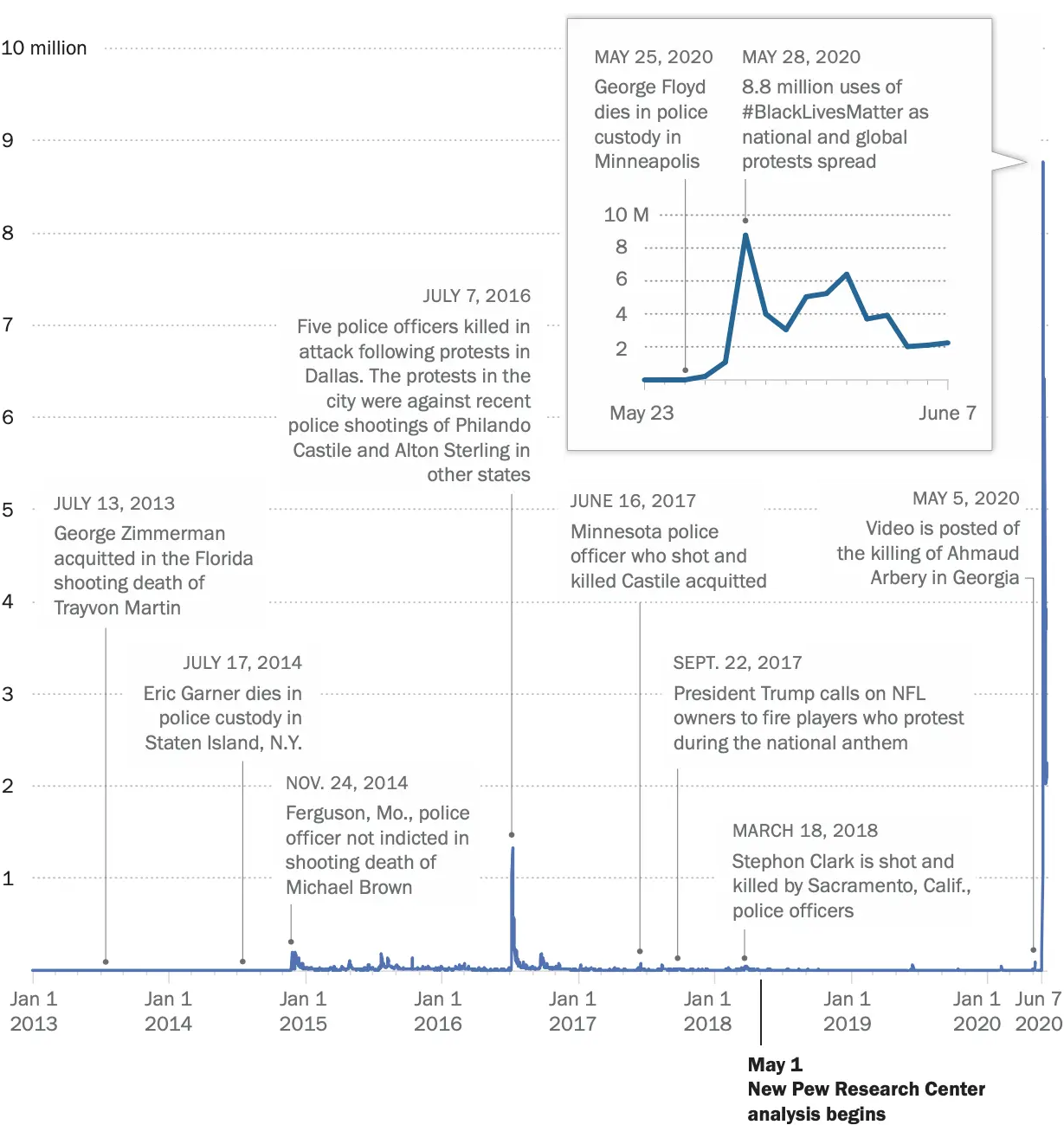}{History of the \#BlackLivesMatter hashtag.}{}

These tweets are collected by monitoring the most important hashtags related to the incident that people have used during the protests and afterwards. 
The data labeling involved monitoring the most important and trustworthy news outlets have published timely reports on the incident, protests and events after it.
The remainder of this paper is organized as follows: Section~\ref{sec:data} introduces the data collection methodology; 
Section~\ref{sec:api} describes the data analysis methods; Section~\ref{sec:baselines} presents baseline results on the dataset; finally, Section~\ref{sec:conc} offers concluding remarks.

\section{Data Collection}
\label{sec:data}
We collected the BLM dataset\footnote{freely available for download at \url{https://github.com/senticnet/BLM}} using various hashtags covering the majority of Tweets about this topic. Such hashtags are listed below:
\begin{itemize}
  \item \texttt{\#BLM}
  \item \texttt{\#BlackLivesMatter}
  \item \texttt{\#JusticeForGeorgeFloyd}
  \item \texttt{\#GeorgeFloyd}
  \item \texttt{\#ICantBreathe}
  \item \texttt{\#PoliceBrutality}
  \item \texttt{\#PeacefulProtest}
\end{itemize}

    After data collection was completed, the tweets window was adjusted between the day of George Floyd incident is happened, 25 May 2020~\cite{reuters_2020}, and 21 August 2020~\cite{abc_george_floyd}, when George Floyd's mural was defaced. The total number of tweets crawled is 16,782,467 (\textit{Table~\ref{tab:hashtag_count}}).
     In Fig.~\ref{fig:total_freqs}, daily frequencies for Tweets are displayed in a time-series manner, showing high values on the first days of incident and a decrease on coming days but the event is still trending 90 days later.
     
    \begin{table}[h]
      \centering
      \begin{tabular}{|c|c|c|c|}
        \hline
        Hashtag & Start Date & End Date & Tweet Count \\
        \hline
         blm & 2020-05-25 & 2020-08-21 & 7,735,737 \\
         \hline
         blacklivesmatter & 2020-05-25 & 2020-08-21 & 1,341,533 \\
         \hline
         justiceforgeorgefloyd & 2020-05-25 & 2020-08-21 & 1,054,728 \\
         \hline
         georgefloyd & 2020-05-25 & 2020-08-21 & 4,545,748 \\
         \hline
         icantbreathe & 2020-05-25 & 2020-08-21 & 251,494 \\
         \hline
         policebrutality & 2020-05-25 & 2020-08-21 & 1,822,548 \\
         \hline
         peacefulprotest & 2020-05-25 & 2020-08-21 & 30,679 \\
         \hline
         \multicolumn{4}{r}{\textit{\textbf{Total:} 16,782,467}} \\
      \end{tabular}
      \caption{Crawled tweet number and date range based on labels.}
      \label{tab:hashtag_count}
    \end{table}

    \makefig{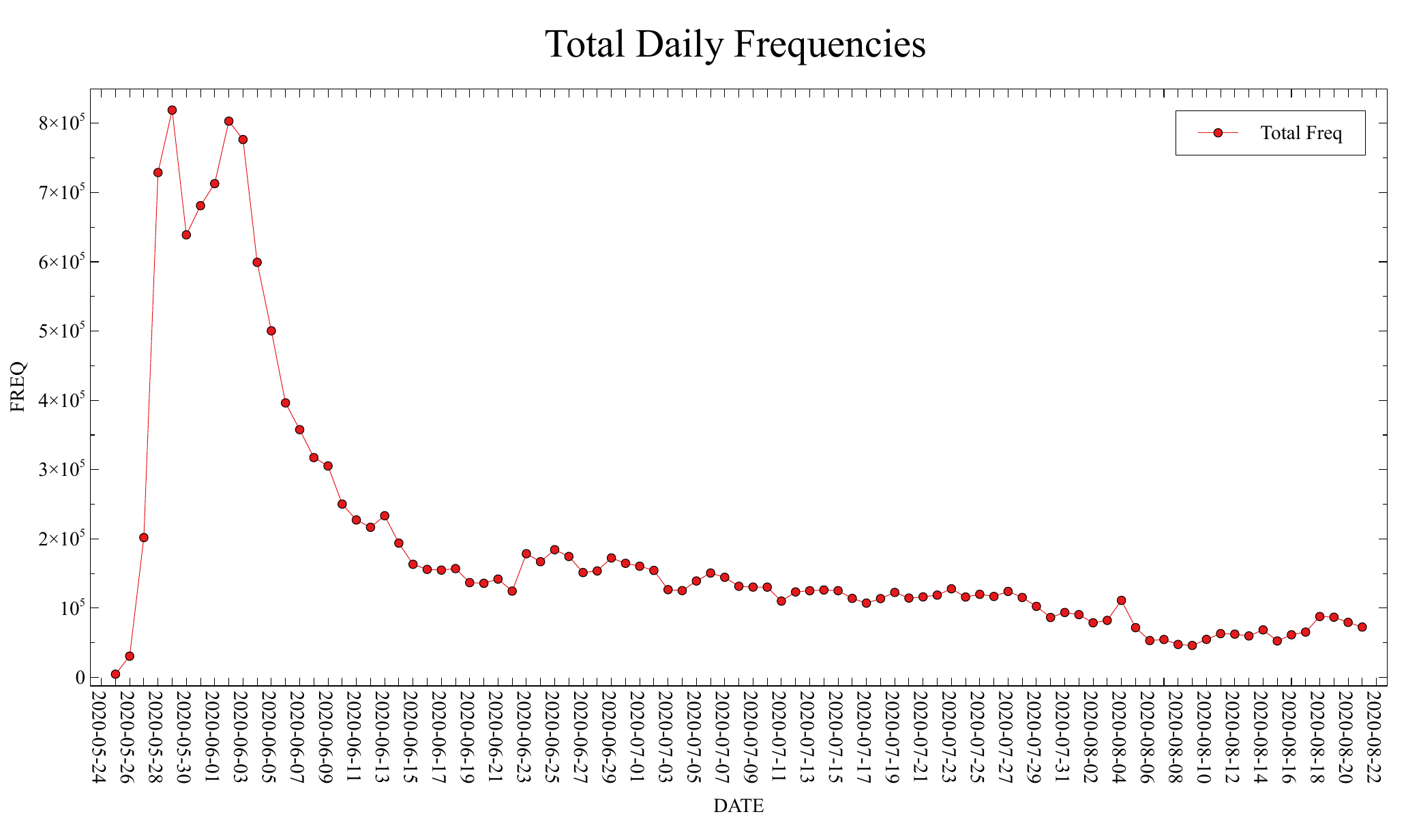}{Total daily tweet numbers.}{}

Data labeling is accomplished using following steps:

\begin{itemize}
  \item[-] Local and international newspapers are examined in order to correctly determine the days of the events.
  \item[-] Websites are also examined in the same way as the newspapers.
  \item[-] After noting down important dates and articles found on newspapers and websites, keywords are extracted from each of them by hand.
\end{itemize}
    
    Fig.~\ref{fig:top_words} represents the top keywords extracted from the dataset on daily basis. We considered these keywords after lemmatization and stop-word removal. We kept the preprocessing minimal not to alter the dataset and its main characteristics.
    \makefig{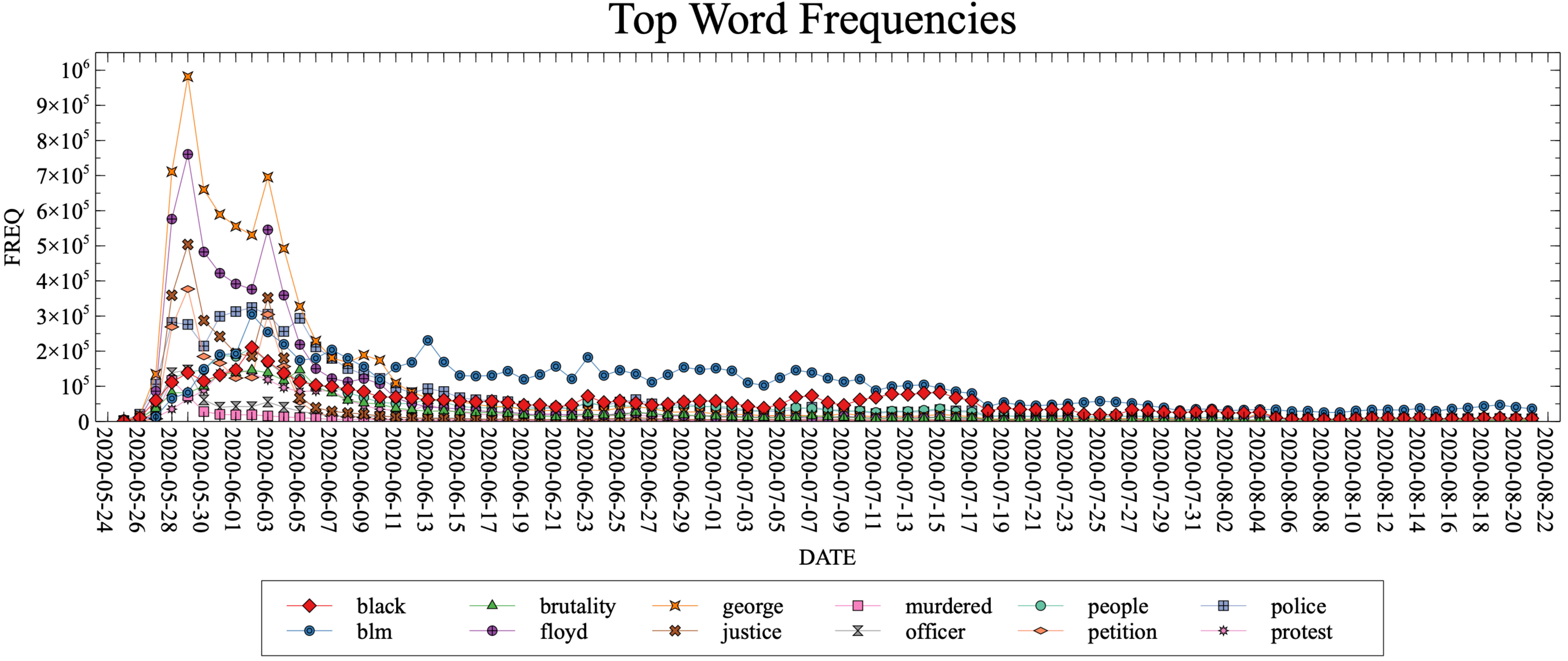}{Frequency analysis of top keywords.}{}
    Some of the extracted keywords are listed below.
      
    \begin{minipage}{.23\textwidth}
     \vspace{10pt} 
    \begin{itemize}
      \item abetting
      \item accident
      \item activist
      \item american
      \item anger
      \item arrest
      \item black
      \item blacklivesmatter
      \item breathe
      \item burial
      \item chaos
       \item governor
      \item ground
      \item guard
      \item head
      \item icantbreathe
      \item injure
      \item justice
      \item killed
      %\item live
      \item manslaughter
      \item murder
    \end{itemize}
    \end{minipage}
    \begin{minipage}{.23\textwidth}
     \vspace{10pt} 

    \begin{itemize}
      \item charged
      \item civil
      \item clash
      \item confederate
      \item crisis
      \item crowd
      \item curfew
      \item death
      \item fire
      \item floyd
      \item george
      \item officer
      \item peace
      \item police
      \item protest
      \item riot
%      \item right
      \item shooting
      \item symbol
      \item trump
      \item washington
      \item whitehouse
    \end{itemize}
    \end{minipage}    
  \vspace{10pt}    
  
     In Fig.~\ref{fig:wordcloud}, most frequent weekly words are visualized after lemmatization and  stopword removal. This weekly view shows trending different keywords on different weeks.
  We labeled 5k sentences from the collected tweets and preprocessed them by removing urls, user mentions, twitter picture links and emojis. We labeled the tweets based on their sentiment and relatedness to BLM. We extracted keywords of the positive and negative tweets. For sentiment analysis, we use four labels: ``1" for the positive tweets, ``2" for the negative tweets, ``3" for the Neutral tweets, and ``4" for the ``No data" tweets (for the tweets that only contain hashtags or mentions). The dataset contains 568 positive tweets, 1972 negative tweets, 1363 neutral tweets, and 648 ``No Data" tweets. For the relatedness to BLM, we used 3 labels: ``0" for not related, ``1" for Related, and ``4" for No data. This dataset contains 648 ``No data" tweets, 560 ``Not related" tweets, and 3343 ``Related" tweets.
  
 \makefig{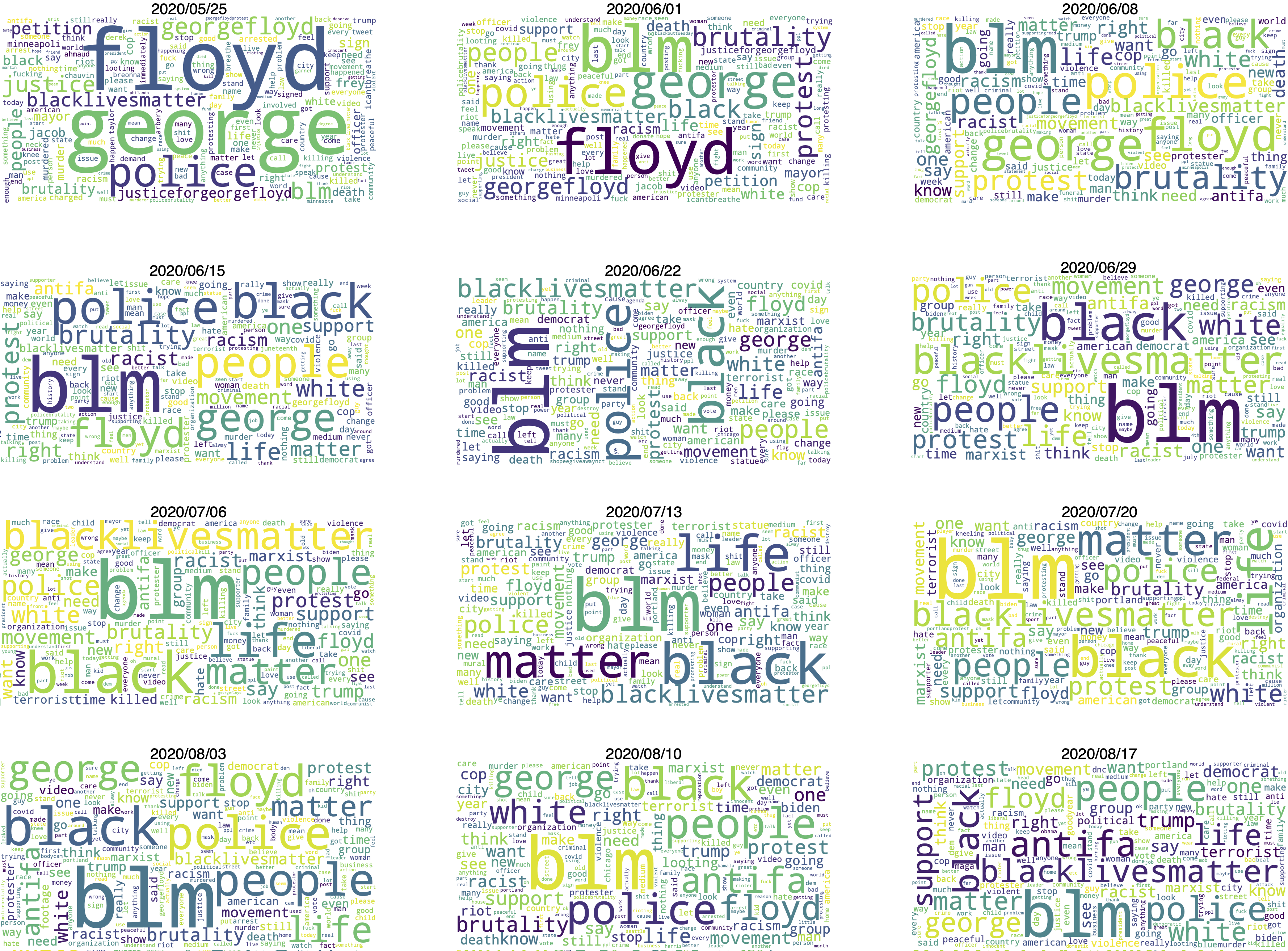}{Weekly WordCloud representations.}{*}

  \begin{minipage}{0.45\textwidth}
    \begin{figure}[H]
      \centering
      \includegraphics[width=\columnwidth]{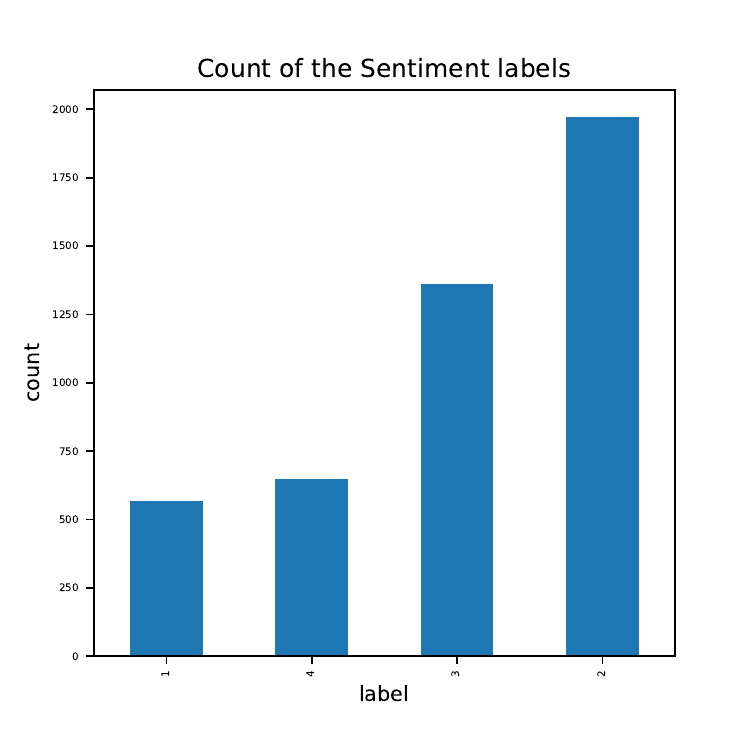}
      \caption{Distribution of the sentiment labels on the labeled BLM dataset.}
      \label{fig:sentimentcount}
    \end{figure}
  \end{minipage}
  \begin{minipage}{0.45\textwidth}
    \begin{figure}[H]
      \centering
      \includegraphics[width=\columnwidth]{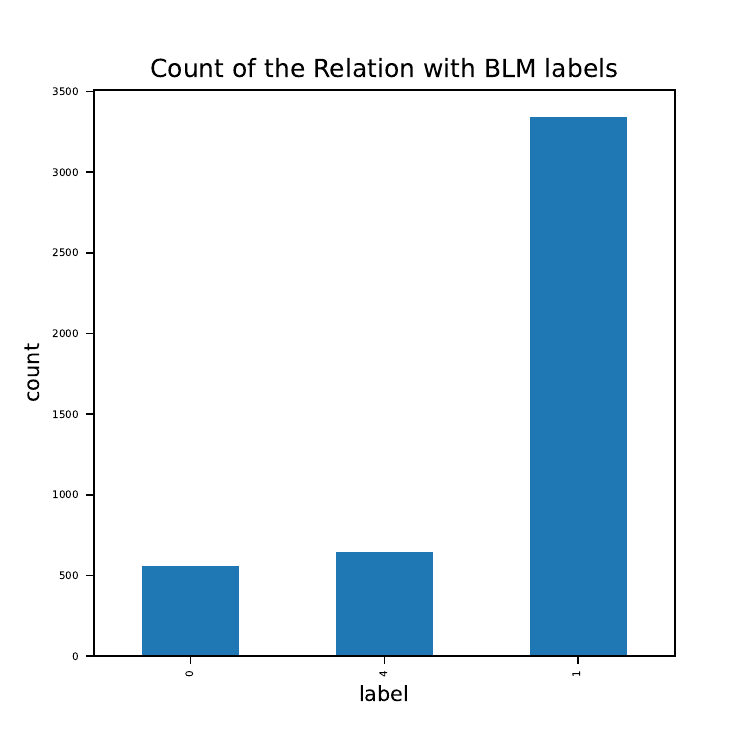}
      \caption{Distribution of the relatedness to BLM  on the labeled BLM dataset.}
      \label{fig:relationcount}
    \end{figure}
  \end{minipage}  
 
 \begin{figure*}[h]
	\centering
	\includegraphics[width=0.91\linewidth]{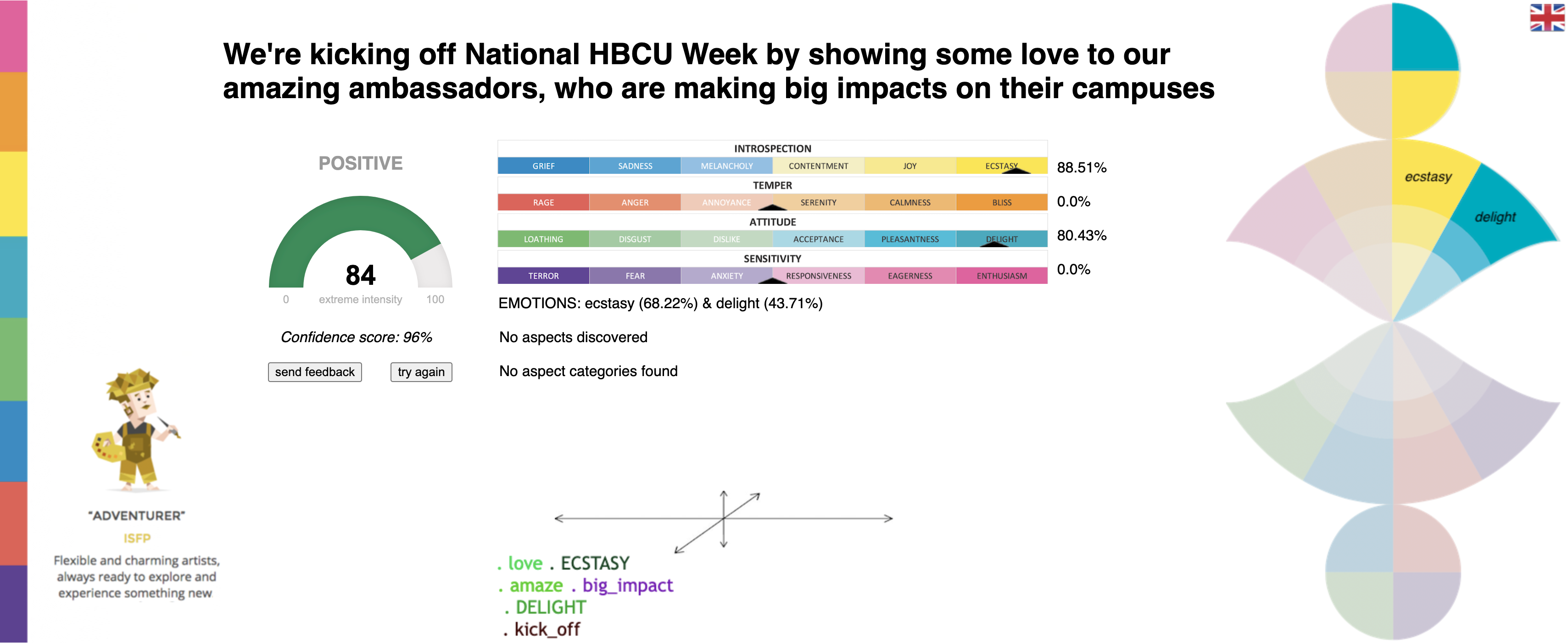}
	\caption{Sentic API user interface sample.}
	\label{fig:ui}
\end{figure*}
 
\section{Data Analysis}
\label{sec:api}
To further enrich the manually labeled data, we use automatic tools for sentiment analysis.
In recent years, sentiment analysis has become increasingly popular for social data analysis~\cite{onesta}. Different AI techniques have been leveraged to improve both accuracy and interpretability of sentiment analysis algorithms, including symbolic AI~\cite{camcomshort,xincog}, subsymbolic AI~\cite{shortcammed,camsta}, and neurosymbolic AI~\cite{valcon,camnt7}. Besides traditional algorithms~\cite{ngulea} focusing on English text, multilingual~\cite{vilbab} and multimodal~\cite{camble} sentiment analysis have also attracted increasing attention recently. Typical applications of sentiment analysis include social network analysis~\cite{cavemb}, finance~\cite{xinint}, and healthcare~\cite{campro}.
In particular, we use Sentic APIs\footnote{freely available at \url{https://sentic.net/api}}, a suite of application programming interfaces, which employ neurosymbolic AI to perform various sentiment analysis tasks in a fully interpretable manner (Fig.~\ref{fig:ui}). 
%All the APIs are based on the sentic computing framework and, hence, leverage an ensemble of symbolic AI (SenticNet) and subsymbolic AI (deep learning). 
A short description of each API and its usage within this work is provided in the next 12 subsections.

\makefig{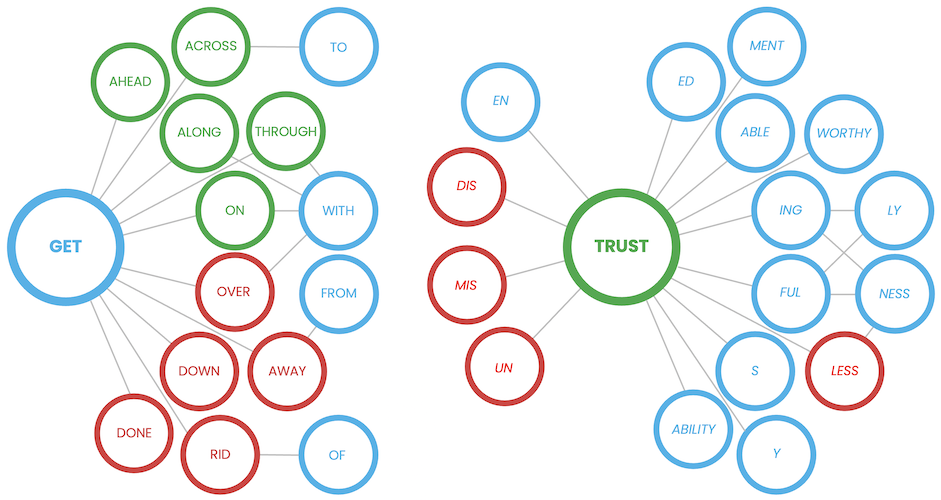}{Sentic Parser graph sample.}{}

\subsection{Concept Parsing}
This API provides access to Sentic Parser~\cite{campar}, a knowledge-specific concept parser based on SenticNet~\cite{camnt7}, which leverages both inflectional and derivational morphology for the efficient extraction and generalization of affective multiword expressions from English text. In particular, Sentic Parser is a hybrid semantic parser that uses an ensemble of constituency and dependency parsing and a mix of stemming and lemmatization to extract `semantic atoms' like pain\_killer, go\_bananas, or get\_along\_with, which would carry different meaning and polarity if broken down into single words (Fig.~\ref{fig:sentic-parser}).
We use the API for extracting words and multiword expressions from tweets in order to better understand what are the key concepts related to BLM as shown in Fig.~\ref{fig:wordcloud}.

\subsection{Subjectivity Detection}
Subjectivity detection is an important NLP task that aims to filter out `factual' content from data, i.e., objective text that does not contain any opinion. 
This API leverages a knowledge-sharing-based multitask learning framework powered by a neural tensor network, which consists of a bilinear tensor layer that links different entity vectors~\cite{satpol}.
We used the API to identify BLM text as either objective (unopinionated) or subjective (opinionated) but also to handle neutrality, that is, text that is opinionated but neither positive nor negative (ambivalent stance towards the opinion target).

\subsection{Polarity Classification}
Once opinionated text is extracted using the \textit{Subjectivity Detection API}, the \textit{Polarity Classification API} further categorizes such text as either positive or negative. 
This is one of the most important APIs we use to understand the stance of tweeters towards BLM.
It leverages an explainable fine-grained multiclass sentiment analysis method~\cite{wanmim}, which involves a multi-level modular structure designed to mimic natural language understanding processes, e.g., ambivalence handling process, sentiment strength handling process, etc. 

\begin{figure}[h]
	\centering
	\includegraphics[width=0.97\linewidth]{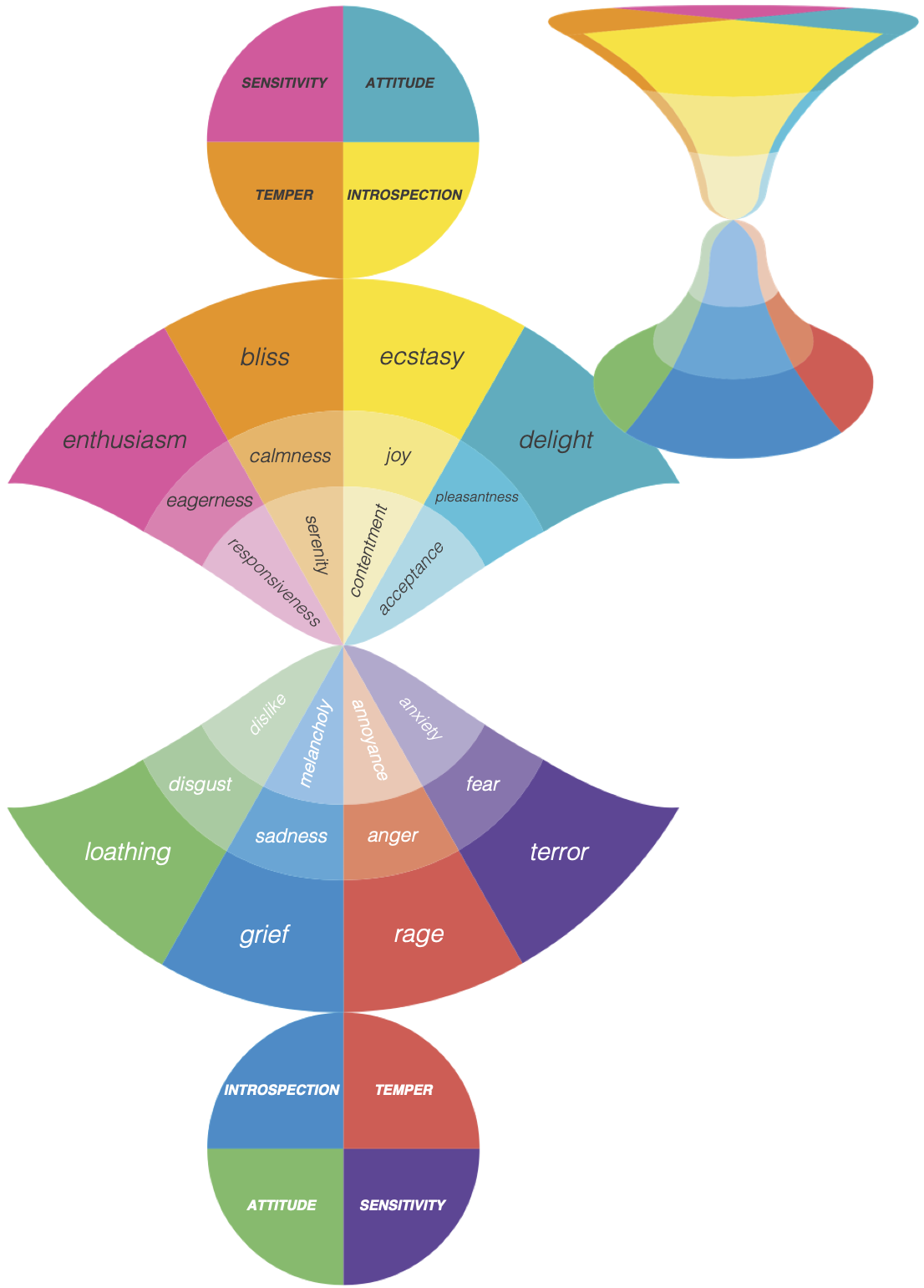}
	\caption{The Hourglass of Emotions.}
	\label{fig:hourglass}
\end{figure}
 
\subsection{Intensity Ranking}
For a finer-grained analysis, we further process the text classified by the \textit{Polarity Classification API} using the \textit{Intensity Ranking API} to infer its degree of negativity (floating-point number between -1 and 0) or positivity (floating-point number between 0 and 1). In particular, the API leverages a stacked ensemble method for predicting sentiment intensity by combining the outputs obtained from several deep learning and classical feature-based models using a multi-layer perceptron network~\cite{akthow}.

\subsection{Emotion Recognition}
This API employs the Hourglass of Emotions~\cite{sushou}, a biologically-inspired and psychologically-motivated emotion categorization model, that represents affective states both through labels and through four independent but concomitant affective dimensions, which can potentially describe the full range of emotional experiences that are rooted in any of us (Fig.~\ref{fig:hourglass}). We use the API to go beyond polarity and intensity by examining what are the specific emotions elicited by BLM in both supporters and opposers of the movement.

\subsection{Aspect Extraction}
This API uses a meta-based self-training method that leverages both symbolic representations and subsymbolic learning for extracting aspects from text. In particular, a teacher model is trained to generate in-domain knowledge (e.g., unlabeled data selection and pseudo-label generation), where the generated pseudo-labels are used by a student model for supervised learning. Then, a meta-weighter is jointly trained with the student model to provide each instance with sub-task-specific weights to coordinate their convergence rates, balancing class labels, and alleviating noise impacts introduced from self-training~\cite{heemet}. 
We use the API to better understand the BLM phenomenon in terms of subtopics or opinion targets. Instead of simply identifying a polarity associated with the whole tweet, the \textit{Aspect Extraction API} deconstructs input text into a series of specific aspects or features to then associate a polarity to each of them. This is particularly useful to process antithetic tweets, e.g., tweets where users list pros and cons of BLM.

\subsection{Personality Prediction}
This API uses a novel hard negative sampling strategy for zero-shot personality trait prediction from text using both OCEAN and MBTI models (Fig.~\ref{fig:personality-prediction}).
In particular, the API leverages an interpretable variational autoencoder sampler, to pair clauses under different relations as positive and hard negative samples,
and a contrastive structured constraint, to disperse the paired samples in a semantic vector space~\cite{zhupae}. 
We use the API to study the different personalities and personas involved in BLM discussions and, hence, better understand the possible drivers of such discussions.

\subsection{Sarcasm Identification}
This API combines commonsense knowledge and semantic similarity detection methods to better detect and process sarcasm in text. 
It also employs a contrastive learning approach with triplet loss to optimize the spatial distribution of sarcastic and non-sarcastic sample features~\cite{yuekno}.
We use the API to understand how much the BLM movement is subject to satire and critique but also to increase the accuracy and reliability of the \textit{Polarity Classification API}.
As sarcasm often involves expressing a sentiment that is opposite to the intended emotion, in fact, it may lead to polarity misclassification and, hence, generate wrong insights and conclusions. 

\makefig{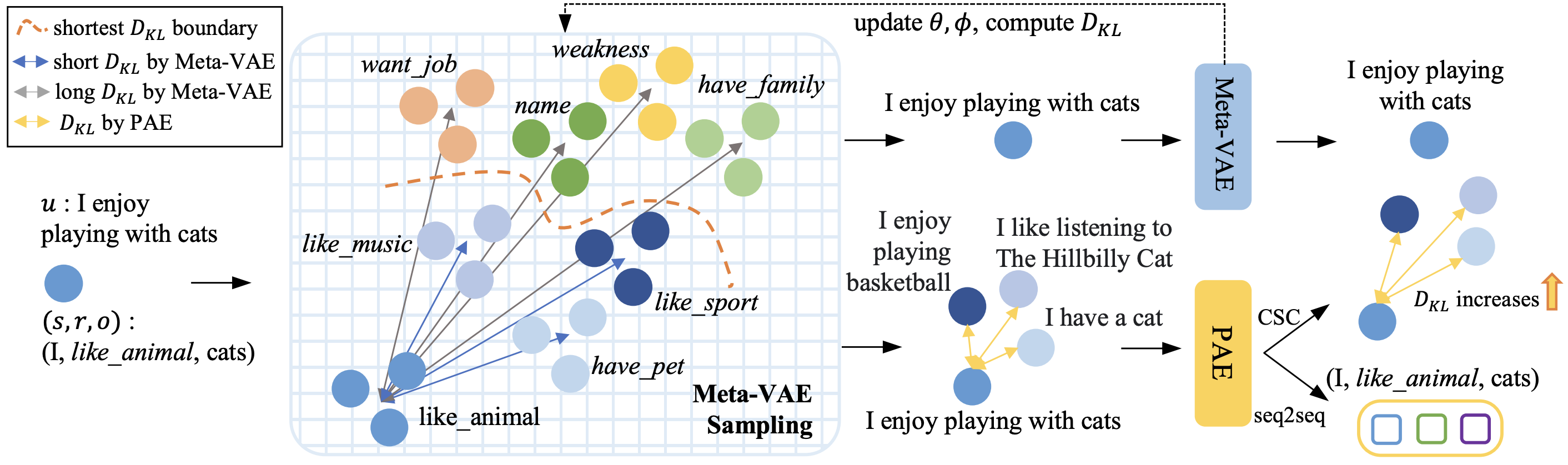}{Personality prediction visualization sample.}{}

%\makefig{absa}{Aspect-based sentiment analysis.}{}

\subsection{Depression Categorization}
This API employs ensemble hybrid learning methods for automated depression categorization.
In particular, the API combines symbolic AI (lexicon-based models) with subsymbolic AI (attention-based deep neural networks) to enhance the overall performance and robustness of depression detection~\cite{ansens}. 
We use it to study different reactions to BLM events by different users, especially those who are negatively affected by them.

\subsection{Toxicity Spotting}
Given the controversy associated with BLM, it is important to measure the different types and intensities of toxicity associated with some of the tweets. 
This API is based on a multichannel convolutional bidirectional gated recurrent unit for detecting toxic comments in a multilabel environment~\cite{kumcom}.
In particular, the API extracts local features with many filters and different kernel sizes to model input words with long term dependency and then integrates multiple channels with a fully connected layer, normalization layer, and an output layer with a sigmoid activation function for predicting multilabel categories such as `obscene', `threat', or `hate' (Fig.~\ref{fig:toxicity-spotting}).

\subsection{Engagement Measurement}
Measuring engagement is important to understand which specific events or topics are more impactful for the BLM movement (Fig.~\ref{fig:blm}).
This API employs a graph-embedding model that fuses heterogeneous data and metadata for the classification of engagement levels.
In particular, the API leverages hybrid fusion methods for combining different types of data in a heterogeneous network by using semantic meta paths to constrain the embeddings~\cite{chapre}.

\makefig{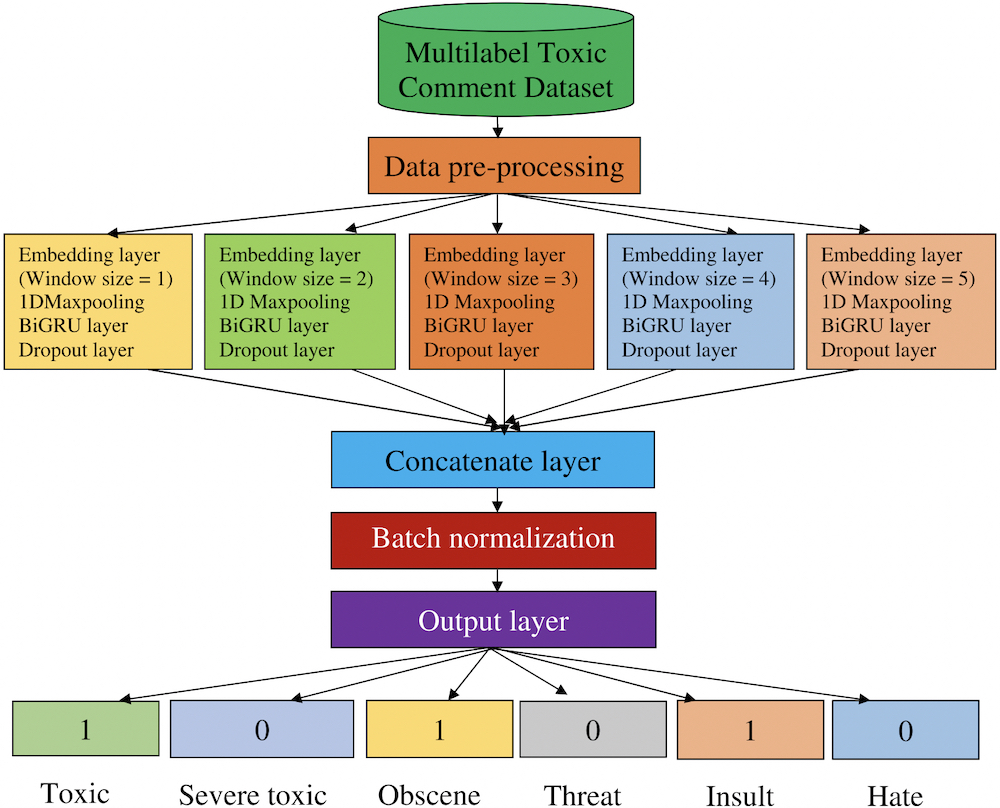}{Toxicity spotting framework.}{}

\subsection{Well-being Assessment}
Besides levels of toxicity and engagement, another important dimension for understanding BLM tweeters is their level of stress.
This API leverages a mix of lexicons, embeddings, and pretrained language models for stress detection from social media texts~\cite{rasstr}.
In particular, the API employs a transformer-based model via transfer learning to capture the nuances of natural language expressions that convey stress in both explicit and implicit manners.

\subsection{API Results}
Through Sentic APIs, we realized that most users used Twitter to share news articles, updates, and announcements related to BLM, including information about protests, rallies, legal developments, and social justice initiatives.
Many also used Twitter to express their support for the BLM movement, share hashtags like \#BlackLivesMatter, and show solidarity with the cause.
Some tweets contained calls to action, encouraging followers to donate to BLM-related organizations, sign petitions, participate in protests, or engage in other forms of activism.
Some users shared personal stories, experiences, or encounters related to racism and discrimination, illustrating the ongoing need for BLM's goals.
%Debates and discussions about BLM's goals, strategies, and effectiveness were also common. 
Finally, Twitter also served as a platform for artists, poets, and writers to share their creative work related to BLM, including visual art, poetry, and essays.
In the figures below, we propose some visualizations of Sentic API results.

    \begin{figure}[H]
      \centering
      \includegraphics[width=\columnwidth]{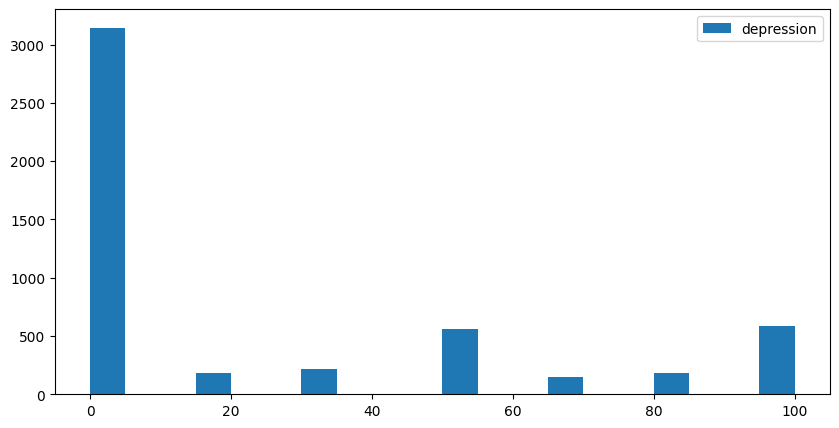}
      \caption{Distribution of depression found in the tweets.}
      \label{fig:depression}
    \end{figure}

    \begin{figure}[H]
      \centering
      \includegraphics[width=\columnwidth]{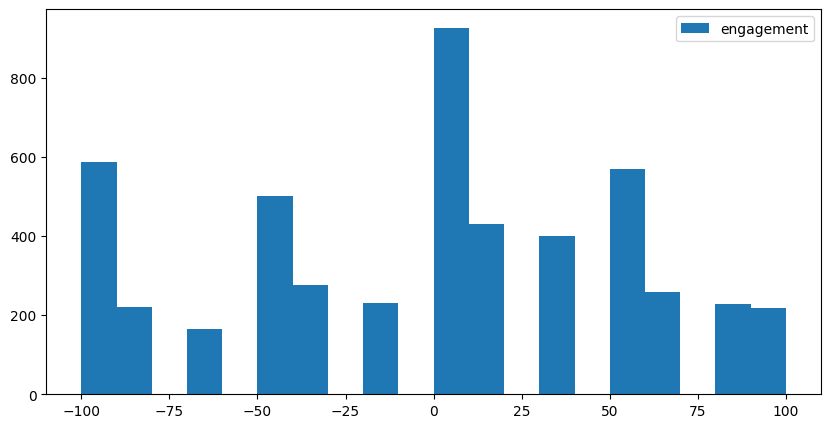}
      \caption{Distribution of engagement found in the tweets.}
      \label{fig:engagement}
    \end{figure}
  
    \begin{table*}[h]
  \centering
  \begin{tabular}{|l|c|c|c|c|c|c|c|}
  \hline
  \multicolumn{8}{|c|}{Score Comparison for Sentiment Classification} \\
  \hline
  \multicolumn{1}{|c|}{Method} & \multicolumn{1}{c|}{Accuracy}&\multicolumn{2}{c|}{Precision} & \multicolumn{2}{c|}{Recall} & \multicolumn{2}{c|}{F-1 Score}\\
  \cline{2-8}
  \multicolumn{1}{|c|}{Name} & - & Macro & Weighted & Macro & Weighted & Macro & Weighted \\
  \hline
  Linear SVC  & 0.62  & 0.61 & 0.62 & 0.53 & 0.62& 0.54 &0.61\\
  
  KMeans&  0.47 & 0.34 & 0.41 & 0.36 & 0.47 & 0.33 & 0.42 \\
  
  Logistic Regression &0.63 & 0.62 & 0.63 & 0.50 & 0.63 & 0.49 & 0.59\\
  \hline
  \end{tabular}
  \vspace{0.2cm}
  \caption{Score Comparison for Sentiment Classification.}
  \label{tab:sentencescores}
  \end{table*}
  
  \begin{figure}[H]
    \centering
    \includegraphics[width=\columnwidth]{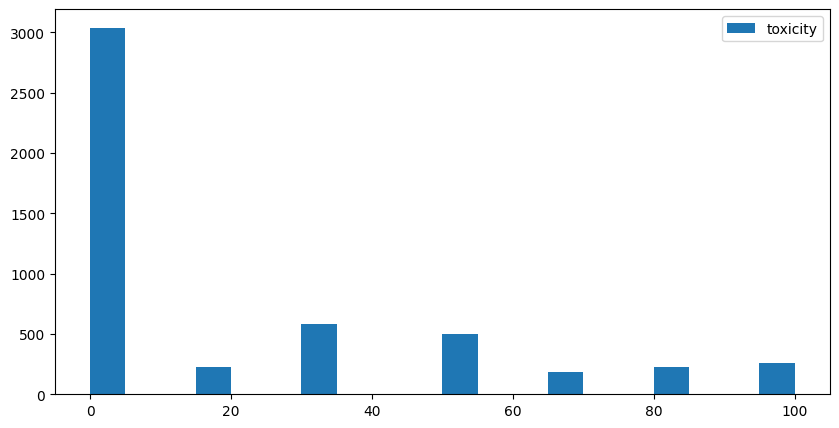}
    \caption{Distribution of toxicity found in the tweets.}
    \label{fig:toxicity}
  \end{figure}

    \begin{table}[h]
    \centering
    \begin{tabular}{|c|c|c|c|c|}
      \hline
       Method& Keyword Count &Precision & Recall & F1-score \\
       \hline 
       TF-IDF & Top 10 & 0.0438 & 0.0250 & 0.0318 \\
       \hline
       LDA & Top 10 & 0.1314 & 0.0750 & 0.0955 \\
       \hline \hline 
       TF-IDF & Top 25 & 0.0949 & 0.0217 & 0.0650 \\
       \hline 
       LDA & Top 25 & 0.2044 & 0.0467 & 0.1400 \\
       \hline \hline 
       TF-IDF & Top 50 & 0.2920 & 0.0333 & 0.1000 \\
       \hline 
       LDA & Top 50 & 0.3066 & 0.0350 & 0.1050 \\
       \hline
    \end{tabular}
    \vspace{0.2cm}
    \caption{Score Comparison for Keyword Extraction.}
    \label{tab:scores}
  \end{table}
  
  Based on Fig.~\ref{fig:engagement}, we can say that the tweets that includes advertisement goes down to -100 range, while the tweets that are actually related to the topic lies right to the 0 point.
  Based on the similarities of distribution in the Fig.~\ref{fig:depression} and Fig.~\ref{fig:toxicity}, we can also see that the majority of the data that we collected mostly convey sadness, rather than aggressiveness, towards the topic.

  \section{Baselines}\label{sec:baselines}
  TF-IDF and LDA extraction methods on daily basis is used to provide baselines for this dataset. Top 10, 25 and 50 extracted keywords is used to check if they exist in the ground truth labels or not. This step can also be considered as a quality check because it ensures that with minimum effort and easier to use methods such as TF-IDF and LDA, results keeps getting better on various top-k analysis.
We used standard metrics as main evaluation anchors are precision, recall and f-measures (as specified in equations~\ref{eq:precision},~\ref{eq:recall} and~\ref{eq:fmeasure}).
  
  \begin{equation}
    Precision = \frac{|\{extracted\_keywords\} \bigcap \{ground\_truth\} |}{|\{ground\_truth\}|}
    \label{eq:precision}
  \end{equation}
  
  \begin{equation}
    Recall = \frac{|\{extracted\_keywords\} \bigcap \{ground\_truth\} |}{|\{extracted\_keywords\}|}
    \label{eq:recall}
  \end{equation}
  
  \begin{equation}
    F1-score = \frac{2 * Precision * Recall}{(Precision + Recall)}
    \label{eq:fmeasure}
  \end{equation}
  
  Results are presented in Table~\ref{tab:scores} for various top-k precision, recall and F1-score. These metrics present an overall improvement between the two baselines, TF-IDF and LDA. According to this table, best obtained results for top-10, 25 and 50 are 0.1314, 0.2044 and 0.3066 for precision.
  Finally, we have also conducted another experiment by using the 17M tweets and a TF-IDF Vectorizer to classify the sentiment scores of the provided tweets. We transformed the tweets from the dataset to vectors by using the trained TF-IDF Vectorizer. By using these vectors, we trained three different sentence classification methods as a baseline. All methods were trained using sklearn with default values.

\section{Conclusion}
\label{sec:conc}
In this work, we have collected a dataset of about 17 million tweets about the George Floyd incident. We have used several machine learning and AI techniques to analyze the collected data and produce several insights.
AI holds immense potential to advance and protect human rights on a global scale. 
We hope this work can foster future investigations on similar topics, such as identifying and combating human rights abuses, enhancing access to justice, and promoting accountability. 

By providing tools for early detection, evidence collection, and advocacy, AI has the potential to significantly contribute to the protection and promotion of human rights worldwide. However, it is crucial to navigate the ethical and privacy implications of AI in this context to ensure that these technologies are used responsibly and in alignment with human rights principles.

\section*{Acknowledgments}
This research is supported by the Agency for Science, Technology and Research (A*STAR) under its AME Programmatic Funding Scheme (Project \#A18A2b0046).

\bibliography{references.bib}
\bibliographystyle{IEEEtran}

\end{document}